\documentclass[runningheads]{llncs}
\usepackage[T1]{fontenc}
\usepackage{cite}
\usepackage{amsmath,amssymb,amsfonts}
\usepackage{algorithmic}
\usepackage{graphicx}
\usepackage{textcomp}
\usepackage{multirow} 
\usepackage{makecell} 
\usepackage{tabularx}

\usepackage{hyperref}
\usepackage{color}

\usepackage{xcolor}
\usepackage{cleveref}
\usepackage{booktabs}
\usepackage{subcaption}
\usepackage{balance}
\usepackage{pgfplots}
\usepackage{float}
\usepackage{xspace}
\usepackage{orcidlink}
\hypersetup{hidelinks}

\def\BibTeX{{\rm B\kern-.05em{\sc i\kern-.025em b}\kern-.08em
    T\kern-.1667em\lower.7ex\hbox{E}\kern-.125emX}}

\crefname{section}{Sec.}{Secs.}
\crefname{table}{Table}{Tables}
\crefname{figure}{Fig.}{Figs.}

\newcommand{\eg}{\emph{e.g.,}~}
\newcommand{\etal}{\emph{et al.}~}
\newcommand{\ie}{\emph{i.e.,}~}
\newcommand{\wrt}{\emph{w.r.t.}~}
\newcommand{\etc}{\emph{etc.}~}

\newcommand{\model}{\textbf{CARe}\xspace}
\newcommand{\dataset}{\textbf{OCTA60}\xspace}
\newcommand{\task}{CMFIR\xspace}

\begin{document}


\title{Cross-modal Fundus Image Registration under Large FoV Disparity}

\titlerunning{Cross-modal Fundus Image Registration under Large FoV Disparity}
%


\author{
Hongyang Li\inst{1}\orcidlink{0009-0002-1412-2861} \and
Junyi Tao\inst{1}\orcidlink{0009-0007-9071-833X} \and
Qijie Wei\inst{1}\orcidlink{0009-0008-6895-5870} \and
Ningzhi Yang\inst{1}\orcidlink{0009-0009-3672-6181} \and
Meng Wang\inst{2} \and
Weihong Yu\inst{2} \and
Xirong Li\thanks{Corresponding author: Xirong Li (xirong@ruc.edu.cn)}\inst{1}\orcidlink{0000-0002-0220-8310}
}
\authorrunning{H. Li et al.}
%
\institute{
Renmin University of China, Beijing, China \and
Peking Union Medical College Hospital, Beijing, China 
\email{https://github.com/ruc-aimc-lab/care}\\
}


\maketitle              

\begin{abstract}
Previous work on cross-modal fundus image registration (\task) assumes small cross-modal Field-of-View (FoV) disparity. By contrast, this paper is targeted at a more challenging scenario with large FoV disparity, to which directly applying current methods fails. We propose \underline{C}rop and \underline{A}lignment for cross-modal fundus image  \underline{Re}gistration(\model), a very simple yet effective method. Specifically, given an OCTA with  smaller FoV as a source image and a wide-field color fundus photograph (wfCFP) as a target image, our \emph{Crop} operation exploits the physiological structure of the retina to crop from the target image a sub-image with its FoV roughly aligned with that of the source. This operation allows us to re-purpose the previous \emph{small-FoV-disparity} oriented methods for subsequent image registration. Moreover, we improve spatial transformation by a double-fitting based \emph{Alignment} module that utilizes the classical RANSAC algorithm and polynomial-based coordinate fitting in a sequential manner. Extensive experiments on a newly developed test set of 60 OCTA-wfCFP pairs  verify the viability of \model for \task. 

\end{abstract}

\section{Introduction} \label{sec:intro}
\begin{figure}[!htb]
    \centering
    \begin{subfigure}[c]{0.8\textwidth}
        \centerline{\includegraphics[width=\linewidth]{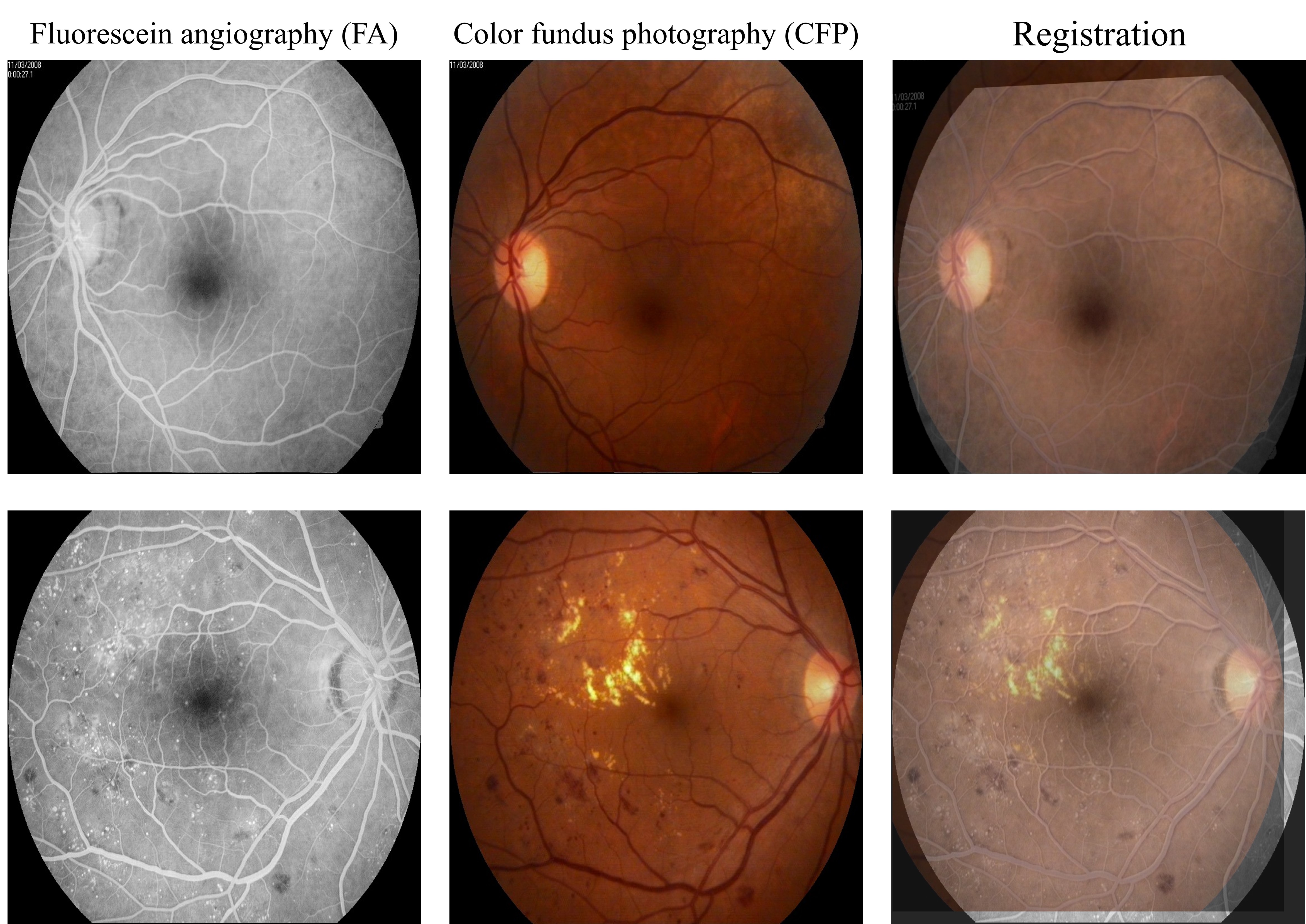}}
	\caption{Small FoV disparity between FA ($\sim$45\textdegree) and CFP ($\sim$45\textdegree)}
	\label{fig:small-fov}
	\end{subfigure}
	\begin{subfigure}[c]{0.8\textwidth}
        \centerline{\includegraphics[width=\linewidth]{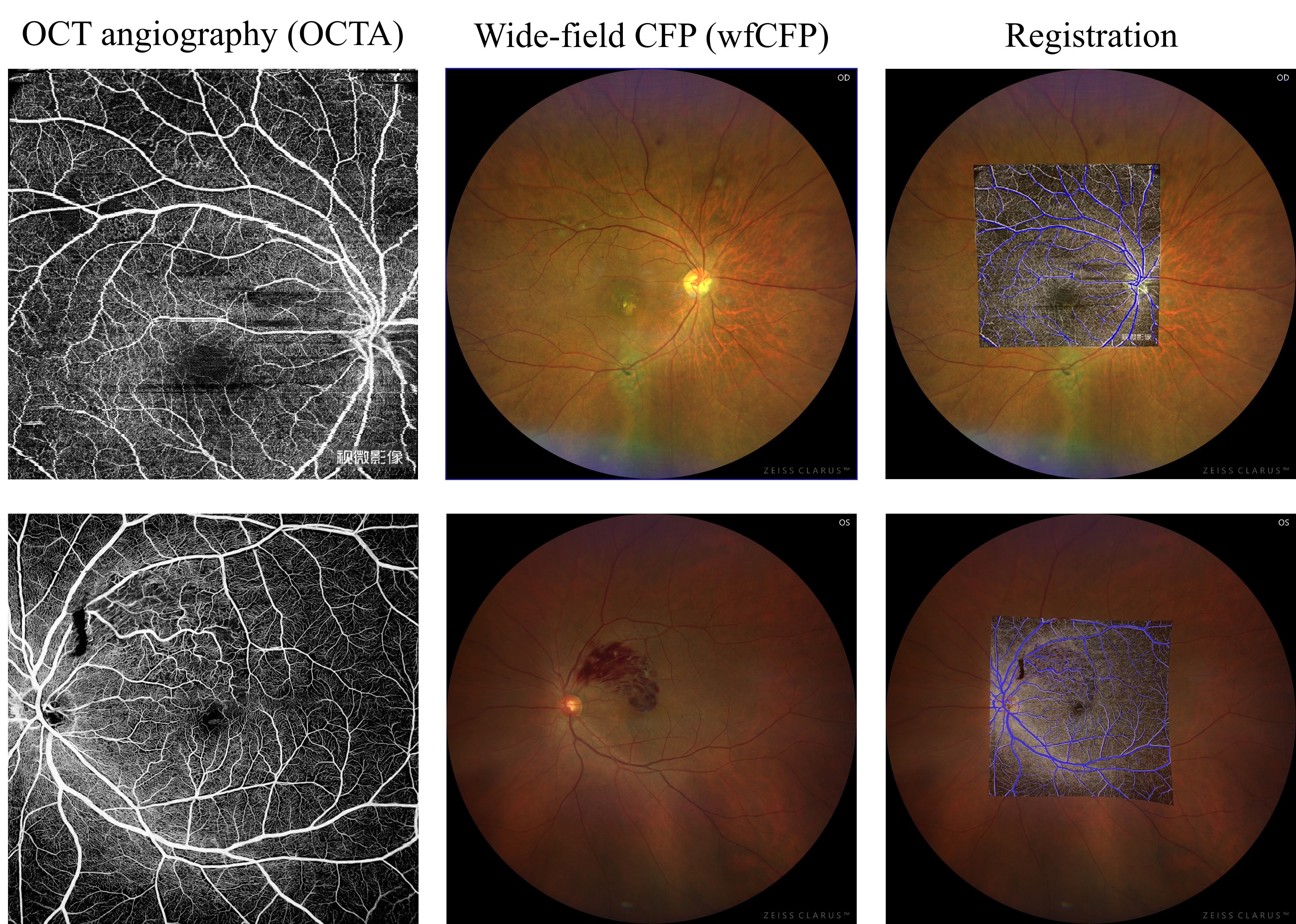}}
	\caption{Large FoV disparity between OCTA ($\sim$40\textdegree) and wfCFP ($\sim$90\textdegree)} 
	\label{fig:large-fov}
	\end{subfigure}
	\hfill
\caption{\textbf{Cross-modal fundus image registration} (CM-FIR)  under (a) small and (b) large FoV disparity. While existing works focus on the former, this paper tackles the latter.}
\label{fig:task}
\end{figure}

This paper aims for \emph{cross-modal} fundus image registration (\task) under large Field-of-View (FoV) disparity, an emerging challenge arising with the development of fundus imaging techniques. More and more retinal lesions can nowadays be visualized in a \emph{noninvasive} manner. Consider for instance non-perfusion area (NPA), a crucial feature of microvascular injury. Previously, NPA had to be identified through invasive fluorescein angiography (FA), see \cref{fig:small-fov}. It can now be identified via noninvasive OCT angiography (OCTA) \cite{b1}. Registering an OCTA image to a wide-field color fundus photograph (wfCFP) of the same eye produces a composite image with enhanced clinical value, see \cref{fig:large-fov}.

\begin{figure}[htb!]
\centering
\includegraphics[width=0.9\textwidth]{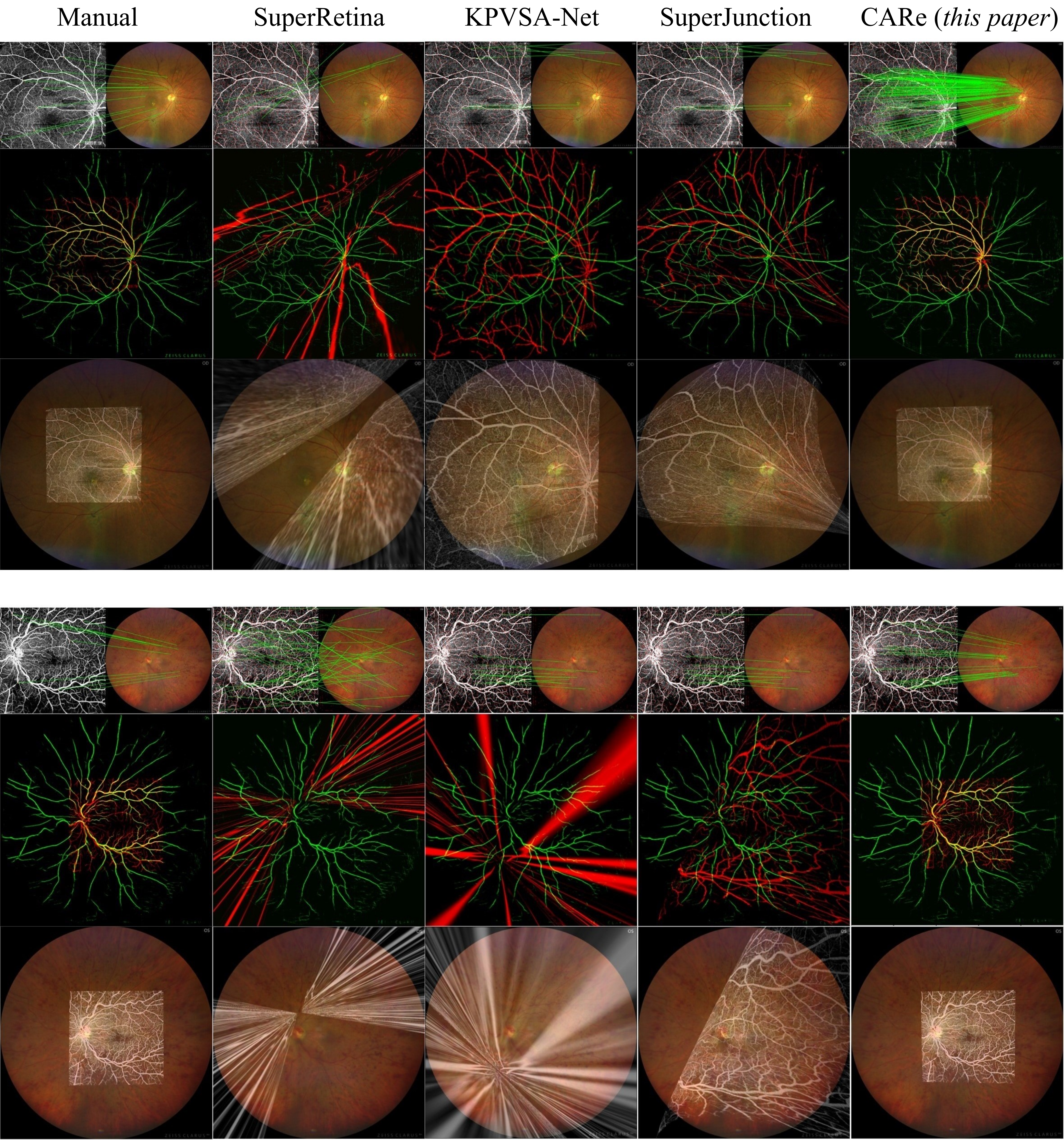}
\caption{\textbf{Visualization of registering a specific OCTA image to a wfCFP image of the same eye}. Vessels in red are from the OCTA, while vessels in green are from the wfCFP. The cross-modal aligned vessels are highlighted in yellow. More yellows mean better alignments. Simply re-training current \emph{small-FoV-disparity} oriented methods (SuperRetina \cite{eccv22-liu}, KPVSA-Net \cite{miccai22-sindel} and SuperJunction \cite{aaai24-wang}) does not work.  Best viewed digitally.}
\label{fig:reg-results}
\end{figure}

(Cross-modal) Fundus image registration has been actively studied \cite{tbe10-chen,icip19-zhang,icassp20-wang,tip22-zhang,miccai22-sindel,eccv22-liu,aaai24-wang}, with feature (or keypoint) based methods as the mainstream solution. In order to register a specific source image \wrt a given target image, a feature-based method typically performs feature or keypoint detection, followed by feature matching to find a correspondence between a set of keypoints. Based on the correspondence, a specific spatial transforming function, with homograpy as a common choice, is then fitted. Much progress has been made, with generic feature detectors \cite{sift04-lowe,superpoint} replaced by fundus-specific alternatives such as SuperRetina \cite{eccv22-liu} and SuperJunction \cite{aaai24-wang}, the brute-force matcher replaced by learnable alternatives like SuperGlue \cite{superglue}, \etc Existing methods typically assume small FoV disparity between the source and target images, see \cref{fig:small-fov}. 
Under large cross-modal FoV disparity, as in the case of OCTA-to-wfCFP, both the feature matching and coordinate fitting processes become erroneous. 
As exempifiled in \cref{fig:reg-results}, simply re-purposing the SOTA methods \cite{eccv22-liu,miccai22-sindel,aaai24-wang} for the new task fails. 

\begin{figure}[!tb]
\centerline{\includegraphics[width=\textwidth]{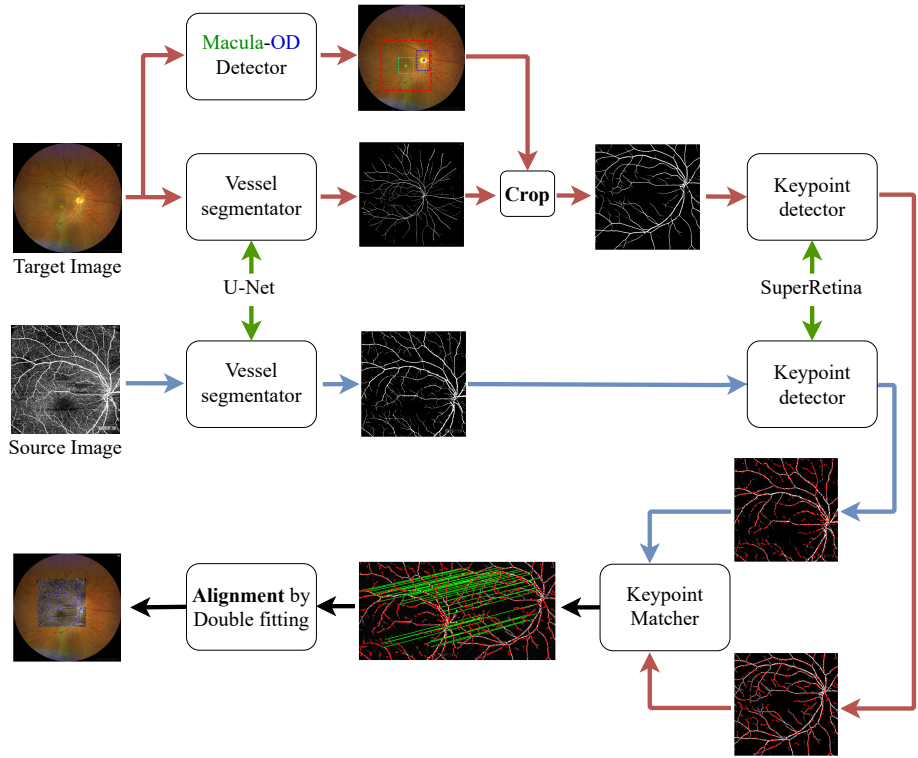}}
\caption{\textbf{Conceptual diagram of the proposed \model method}.
In order to tackle CM-FIR under large FoV disparity, 
our method contains \emph{task-specific} designs as follows: i) U-Net based \emph{unified} vessel segmentation that converts cross-modal fundus images (OCTA and wfCFP) to a unified vessel map representation, ii) \emph{Crop} for coarse FoV alignment achieved by cropping a sub-image from the target image based on jointly detected macula and optic disc (OD), iii) an improved training procedure of SuperRetina for better keypoint matching, and iv) \emph{Align} for source-to-target coordinate transformation by \emph{double fitting}, involving a novel cascaded use of RANSAC and polynomial-based coordinate fitting.}
\label{fig:framework}
\end{figure}

To attack the new challenge, we propose 
\underline{C}rop and \underline{A}lignment for cross-modal fundus image \underline{Re}gistration (\model), see \cref{fig:framework}. Our idea is very simple. Noting the large FoV disparity as the main obstacle, we propose a \texttt{Crop} operation to achieve a rough cross-modal FoV alignment at the first place. 
Despite its simplicity, such an operation allows us to  re-purpose the previous \emph{small-FoV-disparity} oriented methods for subsequent fundus image registration. Furthermore, we 
improve coordinate transformation by a double-fitting that makes a cascaded use of the classical RANSAC algorithm and polynomial-based coordinate fitting.  
To sum up, our major contributions are as follows. 
\begin{itemize}
    \item To the best of our knowledge, we are the first to tackle \task under large FoV disparity. To that end, we build \dataset, a test set of 60 OCT-wfCFP pairs, collected from Outpatient Clinic with images showing varied fundus conditions of real-world patients.
    \item We propose \model, a simple yet effective method that uses a \texttt{Crop} operation, which exploits the physiological structure of wfCFP, to simplify the registration process and an \texttt{Alignment} module to improve the registration accuracy. Moreover, we improve the training procedure of SuperRetina for better keypoint matching in the cross-modal scenario.
    \item Extensive experiments on the \dataset test set verify the superiority of the proposed method over multiple strong baselines \cite{eccv22-liu,miccai22-sindel,aaai24-wang} re-purposed for the new task. 
\end{itemize}




\section{Related Work}

\subsection{Methods for \task}

Current methods for \task are feature-based, extracting discriminative features either from fundus images \cite{icip19-zhang,miccai22-sindel} or from their vascular maps \cite{icassp20-wang,tip22-zhang}, followed by feature matching to establish spatial correspondences between given image pairs. As vascular maps naturally provide a unified representation of cross-modal fundus images, we follow \cite{icassp20-wang,tip22-zhang}, extracting features from the vascular maps. However, different from \cite{icassp20-wang,tip22-zhang} which uses SuperPoint \cite{superpoint}, a generic feature detector and descriptor, we adopt SuperRetina \cite{eccv22-liu} that is specifically developed for fundus image matching. Moreover,  we go one step further by addressing \task under large FoV disparity, a new and more challenging task not considered by the previous works. 

\subsection{Datasets for \task}

To the best of our knowledge, FA-CFP \cite{fa-cfp} is the only public dataset\footnote{\url{https://misp.mui.ac.ir/en/node/1498}}, including 59 pairs of CFP and FA images with small cross-modal FoV disparity, see \cref{fig:small-fov}. As the dataset has no annotation \wrt keypoint correspondence, the registration accuracy of a specific method is indirectly measured by the Dice similarity between the vessel maps of a target image and a registered source image. Our proposed \dataset test set is targeted at the more challenging setting of large cross-modal FoV disparity. Moreover, we provide manually annotated keypoint correspondences per pair for a more comprehensive evaluation.

\section{Proposed 
CARe
Method} \label{sec:method}

Given a pair of cross-modal fundus images captured from the same eye, the goal of \task is to spatially transform the fundus image of smaller FoV, \emph{a.k.a.} the source image, to align with the other image of larger FoV, \emph{a.k.a.} the target image.
More formally, we aim to build a source-to-target coordinate transforming function $F$ that for each pixel positioned by $(u,v)$ in the source image, its counterpart $(x,y)$ in the target image can be well approximated by $F(u,v)$. Our method is feature-based, with $F$ developed based on a correspondence between a set of $m$ keypoints $P=\{((u_i,v_i), (x_i,y_i))\}_{i=1}^m$ obtained by keypoint detection and matching. For keypoint detection and description in a cross-modal manner, we adopt a unified vessel segmentor to convert the cross-modal images to vascular maps. To handle the large cross-modal FoV disparity, we propose a simple yet effective \texttt{Crop} operation,  exploiting the physiological structure of the retina to crop from the target image a region roughly aligned with the source image. Feature matching is then performed on the vessel maps of the source image and the cropped target image. Lastly, for accurate source-to-target coordinate transformation, a double-fitting based \texttt{Alignment} module is developed.

\begin{figure} [!htbp] \centerline{\includegraphics[width=0.9\textwidth]{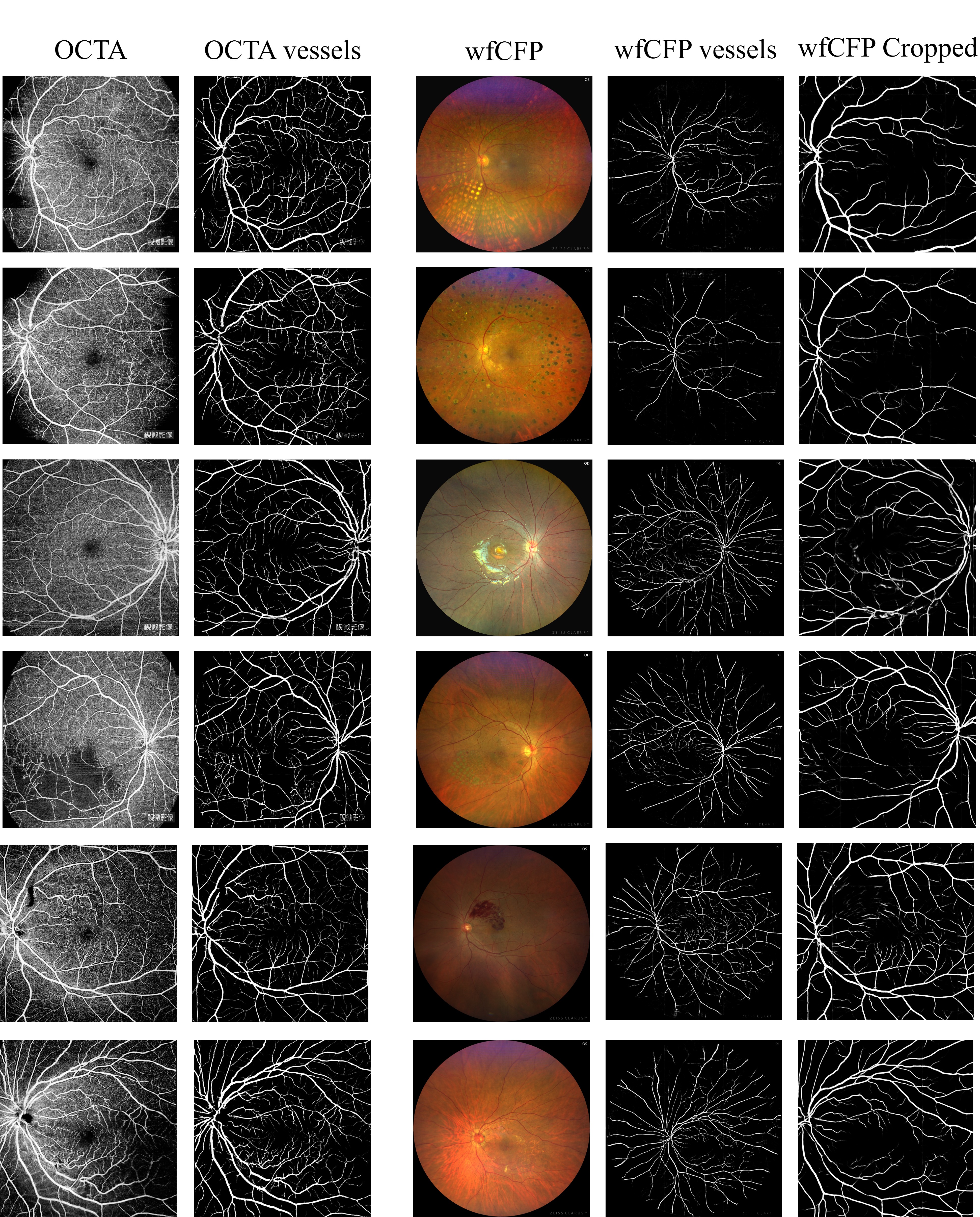}}
    \caption{\textbf{Vessel segmentation and cropping results}. Best viewed digitally.}
    \label{fig:examples}    
\end{figure}

\subsection{Unified Vessel Segmentation} \label{ssec:vessel-seg}

In order to convert the fundus images of distinct modalities to a unified vessel map representation, we train a U-Net \cite{u-net} based vessel segmentation network. Following \cite{icassp25-bdrvs}, our training data is a joint set of public datasets of varied modalities including FIVES for {CFP} \cite{fives}, ROSSA for {OCTA} \cite{rossa}, IOSTAR for scanning laser ophthalmoscopy \cite{iostar}, PRIME-FP20 for ultra-wide-field fundus imaging \cite{prime-fp20} and VAMPIRE for {FA} \cite{vampire}. We empirically find that such a simple solution is sufficient to extract good-quality vessel maps, see \cref{fig:examples}, for keypoint detection and description. An important merit of using the unified vessel segmentor is that our method will be directly applicable to varied modality combinations such as OCTA-CFP and FA-CFP, with no need of combination-specific re-training.

\subsection{\texttt{Crop} for Coarse FoV Alignment} \label{ssec:crop}

As shown in \cref{fig:large-fov}, the source image in this study was obtained by OCTA, which visualizes detailed microvasculature in the macula. Hence, for a coarse FoV alignment between the source and target images, we propose to crop the macular area of the target image. As the macula and the optic disc (OD) are spatially correlated, previous work suggests that jointly detecting the two regions of interest (ROIs) is more accurate and reliable than detecting them alone \cite{mlmi2019-uwf}. In that regard, we train RetinaNet \cite{retinanet}, a widely used one-stage object detection network, on 117 labeled examples. Evaluation on a hold-out test set of 25 images shows that RetinaNet detects both ROIs with mean IoU of around 0.75, sufficiently accurate for our purpose. Based on the auto-located macula and OD, we crop an area roughly corresponding to the posterior pole of the retina. 
As shown in  \cref{fig:framework}, the cropped region is a square centered on the macula, with its side length dynamically determined as twice the distance from the macular center to the outer edge of the detected optic-disc region.
As shown in \cref{fig:examples}, compared to the original target images, the FoVs of the cropped sub-images are much more close to those of the source images.

\subsection{Improved SuperRetina for Keypoint Detection and Matching}

We adopt SuperRetina \cite{eccv22-liu}, an end-to-end network that detects keypoints, \ie \\ crossovers and bifurcations on the vascular tree, and generates their descriptors simultaneously from a given fundus image. Note that SuperRetina, originally developed for uni-modal image registration, takes CFP images as its input. So for input adaptation, we re-train the network on a set of 44 wfCFP vessel maps with manually labeled keypoints. While in theory we shall collect paired images with known spatial correspondence for training, such paired samples are practically generated by applying controlled homography per training image \cite{superpoint}. As such, SuperRetina can be trained in a semi-supervised manner. Given keypoints detected from the vessel maps of the source image and the cropped target image, respectively, we use the classical OpenCV Brute-Force matcher (BFMatcher) to obtain the matched keypoint set $P$.

It is worth pointing out that due to their distinct imaging techniques, OCTA images reveal richer  capillary structures with greater details than their wfCFP counterparts, see \cref{fig:examples}. Recall that SuperRetina is trained exclusively on wfCFP. Such a divergence  introduces a domain gap between keypoint descriptors (or features) learned in the training stage and their counterparts used in the inference stage, and thus makes $P$ suboptimal. 

To bridge the domain gap, we simulate the cross-modal vessel divergence by performing a classical \texttt{opening} operation (an erosion followed by a dilation) on one image per training pair. As a consequence, the processed image loses fine-grained vascular details to some extent. 
Training with such asymmetrically manipulated image pairs allows SuperRetina to better align local features between OCTA and wfCFP images.

\subsection{Fine-grained Alignment by Double Fitting}

For coordinate transformation, previous work typically makes the planar assumption about the fundus images \cite{icassp20-wang,tip22-zhang,eccv22-liu}, hence fitting $F$ by RANSAC-based homography estimation. However, this assumption becomes questionable for large FoV and high-resolution images, \eg 1000$\times$1000 in this work. RANSAC is reliable yet less accurate. The retinal images are two-dimensional projections of the three-dimensional eyeball, making them subject to nonlinear transformations. In contrast, the homography transformation computed by RANSAC is a linear transformation, which can therefore lead to a loss of accuracy. Alternatively, recent work \cite{aaai24-wang} employs polynomial fitting \cite{poly-fitting}. Although accurate, we empirically find that such a fitting strategy is rather sensitive to outliers.

In order to leverage the reliability of RANSAC and the accuracy of polynomial fitting, we propose a very simple yet effective 
double-fitting strategy.
The strategy, dubbed \textbf{RAN-Poly}, first uses RANSAC to remove outliers from $P$, and then fits the following $n$-degree polynomial functions:
\begin{equation} \label{eq:poly}
    \left\{
    \begin{array}{ll}
         x = \sum\limits_{i+j\leq n} a_{ij} u^{i} v^{j} \\
         y = \sum\limits_{i+j\leq n} b_{ij} u^{i} v^{j}, 
    \end{array}
    \right.
\end{equation}
where $a_{i,j}$ and $b_{i,j}$ are polynomial coefficients optimized by the least square method. A larger $n$ means a stronger fitting capability, yet with an increased risk of over-fitting. In the context of uni-modal fundus image registration, Wang \etal \cite{aaai24-wang} empirically show that a quadratic polynomial works the best. We follow their recommendation, using $n=2$ in this work. The validity of this choice is also confirmed by our ablation study. The RAN-Poly double fitting strategy allows us to largely relax the planar assumption, and thus obtain more accurate coordinate transformation.

RAN-Poly is conceptually connected to the bias-variance tradeoff. Polynomial fitting, as its polynomial degree increases, becomes more flexible to fit a training dataset with lower bias, yet there will be greater variance in the model's estimated parameters. By contrast, RANSAC-based homography estimation, with its planar assumption and the fixed number of trainable parameters, has smaller variance yet greater bias. By combining RANSAC and polynomial fitting, RAN-Poly achieves a good bias-variance tradeoff.

\section{Experiments} \label{sec:experiments}

\subsection{Experimental Setup}


\textbf{Test data}. We collect 60 pairs of OCTA and wfCFP images from 60 distinct patients at the outpatient clinic of the Department of Ophthalmology in a state hospital between Feb. 2023 and Dec. 2023.
In particular, OCTA projections of the superficial vascular plexus (SVP) layer were acquired by a 12mm$\times$12mm scan centered on the macula using a SVision VG200 OCT, while wfCFPs were acquired using a ZEISS CLARUS 500 fundus camera. Each image pair has at least 10 manually labeled keypoint correspondences as ground truth. All images are resized to 1000$\times$1000 in advance. They shows varied fundus conditions of real-world patients, challenging the robustness of the proposed method.  This study is complied with the Declaration of Helsinki. 
We term the test set \textbf{\dataset}.

\textbf{Performance metrics}. 
Following \cite{eccv22-liu}, we report the failed / inaccurate / acceptable rate. A registration is considered acceptable if the median Euclidean error (MEE) and the maximum Euclidean error (MAE) between the mapped keypoints and the ground truth are less than 20 and 50 pixels, respectively. We also report AUC with the MEE threshold ranging from 1 to 25. For a more in-depth analysis, we report the average number of keypoint matches per OCTA-wfCFP pair. A match is considered \emph{acceptable} if the matched keypoint is within 20 pixels to the corresponding reference point.
In addition, we report soft Dice Coefficient Dice$_s$ \cite{icip19-zhang}, reflecting the overlap between the vessel maps of the target image and the registered source image.

\textbf{Implementation}.
Training details of the networks used for vessel segmentation, macula / OD detection and keypoint detection  are listed in Table \ref{tab:detail}. All experiments were conducted on an NVIDIA 2080ti GPU with the following software environment: Ubuntu 18.04, CUDA 11.7 and PyTorch 1.21.

\begin{table}[htb]

\renewcommand\arraystretch{1.2}
\centering
\caption{\textbf{Training details of the three deep networks used in the proposed method}: U-Net for vessel segmentation, Retina-Net for macula / OD detection and SuperRetina for feature detection and description.}

\begin{tabular}{@{}lcrrr@{}}
\toprule
\textbf{Network}  & \textbf{Optimizer} & \textbf{Epochs} & \textbf{Learning rate} & \textbf{Batch size}\\ \hline
U-Net  & SGD & 100 & 1e-3 & 5\\ 
RetinaNet & Adam & 150 & 1e-5 & 1\\ 
SuperRetina & Adam & 150 & 1e-3 & 4 \\ \hline
\end{tabular}
\label{tab:detail}

\end{table}

\textbf{Baselines}. We compare with the following SOTA feature-based methods, \ie SuperRetina \cite{eccv22-liu}, KPVSA-Net \cite{miccai22-sindel} and SuperJunction \cite{aaai24-wang}. See \cref{tab:main} for their choices of keypoint matcher and transformation fitting.  For a fair comparison we re-train their modules on our training data whenever applicable. In particular, the same vessel segmentor and keypoint detector are used. Comparison with detector-free methods \cite{iccv23-geoformer} is reserved for future work.

\subsection{Comparison with Baseline Methods}

\textbf{Results on \dataset}. As shown in \cref{tab:main} and \cref{fig:reg-results}, all the  baselines fail to generate acceptable registration results, showing their ineffectiveness in handling the large inter-modal FoV disparity. Note that SuperGlue tends to yield more matches, albeit incorrect, than BFMatcher. Therefore, the failed rate of \cite{miccai22-sindel,aaai24-wang}, which use SuperGlue, is much lower than that of \cite{eccv22-liu} which uses BFMatcher. Our method clearly outperforms the baselines in terms of all performance metrics.

\begin{table}[!htb]
\renewcommand{\arraystretch}{1.2}
\caption{\textbf{Results on \dataset}. 
DLT: Direct linear transformation. 
Poly: Polynomial fitting. RAN: RANSAC.  }\label{tab:main}
\scalebox{0.85}{
\begin{tabular}{@{}l|c|c|r|r|r|r|r@{}}
\toprule
\textbf{Method} & \textbf{Matcher} & \textbf{Fitting} & \makecell[c]{Failed[\%]$\downarrow$} & \makecell[c]{Inaccurate[\%]$\downarrow$} & \makecell[c]{Acceptable[\%]$\uparrow$} & \makecell[c]{AUC$\uparrow$} & Dice$_s$$ \uparrow$\\
\hline
SuperRetina\cite{eccv22-liu}	& BFMatcher & RANSAC & 78.33 & 20.00 & 1.67 & 0.016 & 0.082\\
KPVSA-Net\cite{miccai22-sindel} & SuperGlue & DLT & 8.33 & 91.67 & 0 & 0 & 0.077\\
SuperJunction\cite{aaai24-wang} & SuperGlue & Poly & 23.33 & 76.67 & 0 & 0 & 0.076\\
\hline
\model & \multirow{5}*{BFMatcher} & RAN-Poly & \textbf{0.00} & \textbf{1.67} & \textbf{98.33} & \textbf{0.920} & \textbf{0.295}\\
\emph{w/o} \texttt{Crop} & ~ & RAN-Poly & 55.00 & 43.33 & 1.67 & 0.019 & 0.051\\
\emph{w/o} \texttt{Opening} & ~ & RAN-Poly & 3.33 & 3.33 & 93.33 & 0.896 & 0.282\\
\emph{w/o} RANSAC & ~ & Poly & 0.00 & 31.67 & 68.33 & 0.596 & 0.219\\
\emph{w/o} Poly & ~ & RANSAC & 0.00 & 3.33 & 96.67 & 0.912 & 0.289\\
\bottomrule
\end{tabular}
}
\end{table}

For a better understanding, for each method, we report in \cref{tab:matches} the average number of keypoint matches per OCTA-wfCFP pair. A match is considered \emph{acceptable} if the matched keypoint is within 20 pixels to the corresponding reference point. Facing large cross-modal FoV disparity, the existing methods have much fewer matches with much lower acceptable match rate. Hence, the fact that existing methods struggle with large FoV disparity can be largely attributed to their incapability to find sufficient and acceptable keypoint matches for the subsequent transformation fitting.

\begin{table}[htb!]
\setlength{\abovecaptionskip}{0.cm}
\setlength{\belowcaptionskip}{-0.cm}

\renewcommand\arraystretch{1.2}
\centering

\caption{\textbf{Averaged number of keypoint matches per OCTA-wfCFP pair}.}
\scalebox{1.0}{

\begin{tabular}{@{}lrr@{}}
\toprule
\textbf{Method}  & \makecell[c]{\textbf{\#Matches} $\uparrow$} & \makecell[c]{\textbf{Acceptable-match rate} $\uparrow$} \\
\hline
\makecell[l]{SuperRetina \cite{eccv22-liu} } & 20.8 & 0.152\\ 
\makecell[l]{KPVSA-Net \cite{miccai22-sindel}}  & 35.3 & 0.008\\ 
\makecell[l]{SuperJunction \cite{aaai24-wang}}  & 35.3 & 0.008\\ 
\model  & \textbf{149.46} & \textbf{0.978}\\
\bottomrule
\end{tabular}

}

\label{tab:matches}

\end{table}

\textbf{Results on FA-CFP}. To check how our \texttt{Alignment} module\footnote{Here we omit the \texttt{Crop} operation, which makes no practical difference for CFP with relatively small FoV as in the FA-CFP dataset.} works in the traditional setting (of small cross-modal FoV disparity), we  evaluate the module on the public FA-CFP dataset \cite{fa-cfp}. We report Dice$_s$ only, as the FA-CFP dataset has no keypoint correspondence,
which makes it unfeasible to calculate metrics other than Dice$_s$.
As \cref{tab:cf-fa} shows, the \texttt{Alignment} module alone again surpasses the baseline methods even when the cross-modal FoV disparity is relatively small.

\begin{table}[!tb]
\centering
\renewcommand{\arraystretch}{1.2}
\caption{\textbf{Results on the FA-CFP testset}.
}\label{tab:cf-fa}
\scalebox{1}{
\begin{tabular}{@{}lr@{}}
\toprule
\textbf{Method} & Dice$_s$ \\
\midrule
SuperRetina \cite{eccv22-liu} & 0.530\\
KPVSA-Net \cite{miccai22-sindel} & 0.319\\
SuperJunction \cite{aaai24-wang} & 0.243\\
\model & \textbf{0.556} \\
\bottomrule
\end{tabular}
}
\end{table}

\subsection{Ablation Study}

\textbf{The importance of \texttt{Crop}}. As \cref{tab:main} shows, using the \texttt{Crop} operation or not  has a decisive impact on the performance.

\textbf{Whether \texttt{Crop} helps the baselines}? The answer is yes, see \cref{tab:crop-baselines}, with the largest improvement made for \cite{eccv22-liu}.
Still, our method is better than the much improved baseline, showing the superiority of RAN-Poly to RANSAC.

\begin{table}[!ht]
\centering
\renewcommand{\arraystretch}{1.2}
\caption{\textbf{Performance of the baseline methods with (+) and without (-) our \texttt{Crop} operation}. Test set: \dataset.} \label{tab:crop-baselines}
\scalebox{1}{
\begin{tabular}{@{}lcrrr@{}}
\toprule
\textbf{Method} & \texttt{Crop}? & Acceptable[\%]$\uparrow$ & AUC$\uparrow$ & Dice$_s$$\uparrow$\\
\hline
\multirow{2}*{SuperRetina \cite{eccv22-liu}} & - & 0 & 0 & 0.081\\
~ & + & \textbf{91.67} & \textbf{0.861} & \textbf{0.269}\\
\hline
\multirow{2}*{KPVSA-Net \cite{miccai22-sindel}} & - & 0 & 0 & 0.078\\
~ & + & \textbf{60.00} & \textbf{0.587} & \textbf{0.140}\\
\hline
\multirow{2}*{SuperJunction \cite{aaai24-wang}} & - & 0 & 0 & 0.079\\
~ & + & \textbf{36.67} & \textbf{0.248} & \textbf{0.116}\\
\bottomrule
\end{tabular}
}
\end{table}

\textbf{Impact of the \texttt{opening} operation}. Without the opening operation, AUC drops from 0.9833 to 0.9333 (\cref{tab:main}). Hence, adding the operation improves SuperRetina-based keypoint matching. 

\textbf{The necessity of RAN-Poly}. Replacing RAN-Poly either by polynomial fitting or by RANSAC causes performance degeneration, see the last two rows in \cref{tab:main}. The necessity of our proposed RAN-Poly strategy is thus justified. 

\begin{figure} [!htb]
    \centerline{\includegraphics[width=0.7\textwidth]{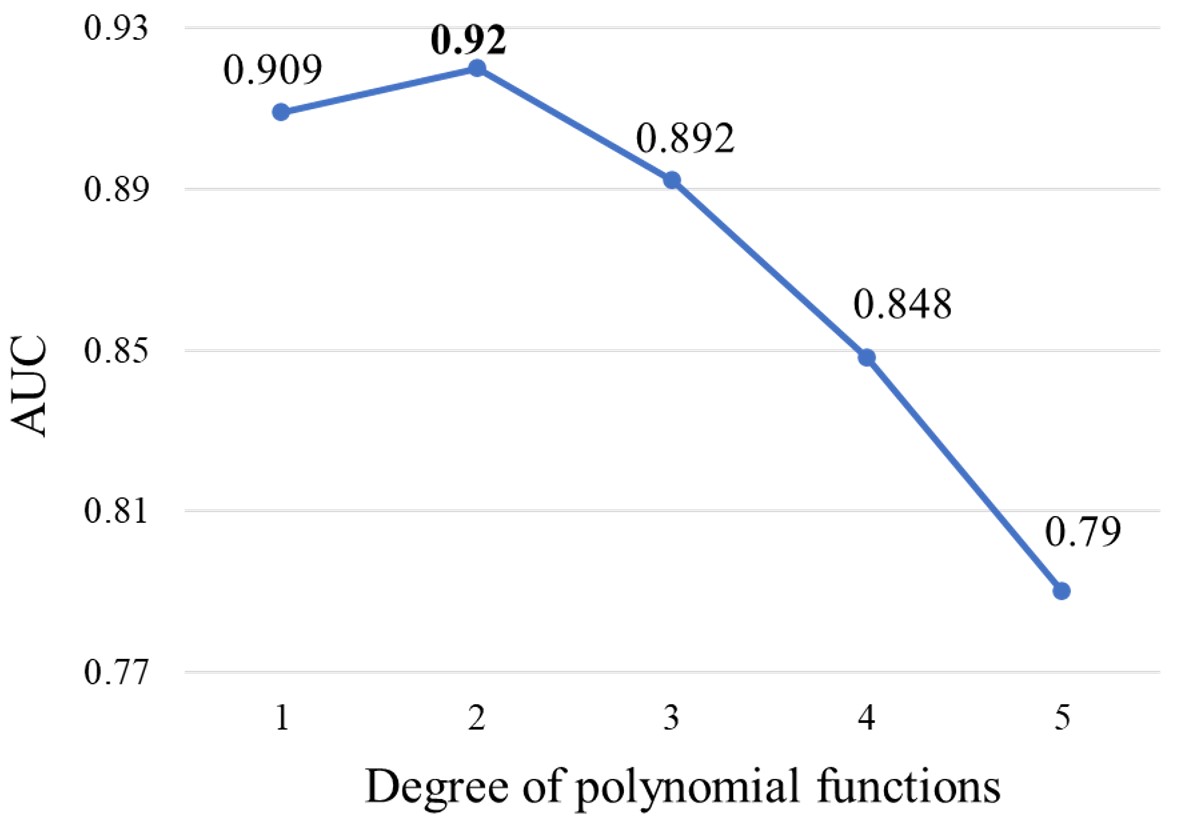}}
    \caption{\textbf{Performance of \model with different degrees of polynomial functions used for polynomial fitting}. The quadratic polynomial is the best.}
    \label{fig:degree}
    
\end{figure}

\textbf{Choice of the polynomial degree}. \cref{fig:degree} shows the performance curve  of the proposed method with different degrees of polynomial functions. The second-degree polynomial strikes the best balance between improving the fitting accuracy and reducing the risk of over-fitting.

\section{Conclusions} \label{sec:conclusion}

This paper tackles an emerging challenge of cross-modal fundus image registration (\task) with large FoV disparity. Experiments on the newly developed \dataset test set allow us to draw conclusions as follows. Directly re-purposing existing small-FoV-disparity oriented methods does not work. Coarse FoV alignment by the proposed \texttt{Crop} operation is crucial. The SOTA methods can be much improved by this operation, though still less effective than the proposed  \model
method. For the estimation of the spatial transforming function, the double-fitting strategy, \ie RAN-Poly, is better than using RANSAC or polynomial fitting alone. While targeted at large cross-modal FoV disparity, our method also works well in the conventional small-FoV-disparity scenario.  With \model and \dataset, we establish a new 
baseline for \task.

\subsubsection{\ackname} This research was supported by National Natural Science Foundation of China (62576348, 62172420) and Beijing Natural Science Foundation (L254039).

\bibliographystyle{splncs04}
\bibliography{C-A-bibliography}
%





\end{document}